\documentclass[conference]{IEEEtran}
\IEEEoverridecommandlockouts
\usepackage{cite}
\usepackage{amsmath,amssymb,amsfonts}
\usepackage{algorithm}
\usepackage{algorithmic}
\usepackage{graphicx}
\usepackage{textcomp}
\usepackage{xcolor}
\usepackage[T1]{fontenc}
\usepackage[utf8]{inputenc}
\usepackage{microtype}
\usepackage{inconsolata}

\providecommand{\citep}[1]{\cite{#1}}

\def\BibTeX{{\rm B\kern-.05em{\sc i\kern-.025em b}\kern-.08em
    T\kern-.1667em\lower.7ex\hbox{E}\kern-.125emX}}

\begin{document}

\title{Latent Visual Contrastive Reasoning with Multi-Branch Exploration}

\author{
\IEEEauthorblockN{
Kegeng Tang\textsuperscript{1},
Jingbo Wang\textsuperscript{1},
Yangming Zhang\textsuperscript{2},
Huanshu Zhang\textsuperscript{3},
Shaogang Ren\textsuperscript{1},
and Zihao Wang\textsuperscript{1}*
}
\IEEEauthorblockA{
\textsuperscript{1}Department of Computer Science \& Engineering,
University of Tennessee at Chattanooga, Chattanooga, TN, USA\\
\textsuperscript{2}Department of Computer Science,
University of Texas at Arlington, Arlington, TX, USA\\
\textsuperscript{3}Department of Electrical Engineering,
The Pennsylvania State University, University Park, PA, USA\\
\{hcw575, ydk297, sren9, zihao.wang\}@tennessee.edu,
yangming.zhang@uta.edu,
hpz5226@psu.edu\\
*Corresponding author: zihao.wang@tennessee.edu
}
}

\maketitle

\begin{abstract}
Multimodal large language models have made substantial progress in visual question answering, but complex visual reasoning often requires locating and comparing multiple pieces of question-relevant visual evidence. Existing methods that move intermediate reasoning into the visual latent space usually follow a single linear reasoning trajectory, which can be limiting when multiple visual hypotheses or competing evidence regions are relevant. We propose Latent Visual Contrastive Reasoning, a branch-aware latent reasoning framework that generates multiple question-conditioned latent visual branches as soft hypotheses over image tokens. To train these branches, our method uses bounding-box evidence annotations to build target evidence representations and introduces a soft contrastive branch-evidence alignment objective. This objective encourages evidence-relevant branches to align with annotated visual regions while contrasting them against negative evidence. We instantiate our method on Qwen2.5-VL backbones and train it with standard supervised fine-tuning. Experiments on perception-intensive visual reasoning benchmarks show that the proposed method achieves overall improvements over the corresponding backbone and single-path latent reasoning baselines at both 3B and 7B scales.
\end{abstract}

\section{Introduction}

Multimodal large language models (MLLMs) have advanced vision-language understanding by connecting visual encoders with large language models \citep{alayrac2022flamingo}. 
This paradigm has been further strengthened by frozen vision-language bootstrapping, visual instruction tuning, and stronger general-purpose multimodal modeling \citep{li2023blip2,liu2023llava,bai2025qwen25vl}. 
Despite these advances, complex visual question answering requires more than recognizing objects or directly mapping an image-question pair to an answer. 
Models must identify question-relevant evidence, compare candidate regions, and reason over fine-grained visual details \citep{antol2015vqa,hudson2019gqa}. 

A common way to improve reasoning is to elicit chain-of-thought (CoT) rationales in the language space \citep{wei2022chain}. 
For multimodal reasoning, this encourages the model to decompose a visual question into intermediate textual steps before answering \citep{zhang2023multimodalcot}. 
Although textual reasoning can improve interpretability and accuracy, it remains fundamentally language-centric. 
The image is typically encoded once and treated as a static condition, while intermediate thoughts are represented as discrete text tokens rather than visual evidence.

To better incorporate visual information, recent methods augment MLLMs with explicit visual operations, such as region selection, cropping, zooming, tool invocation, or program execution. 
Visual programming methods compose perception modules and symbolic programs for compositional visual reasoning \citep{gupta2022visprog}. 
Tool-augmented approaches allow language models to invoke external visual tools or execute programs during inference \citep{suris2023vipergpt,lu2023chameleon}. 
VISUAL COT further provides visual chain-of-thought annotations with bounding-box evidence for multimodal reasoning \citep{shao2024visualcot}. 
These approaches enable models to dynamically focus on image regions and re-inject localized observations into later reasoning steps. 
However, they often depend on external modules, hand-designed operations, or repeated image re-encoding, and the final reasoning trajectory is still largely expressed in natural language.

Latent Visual Reasoning (LVR) offers a promising alternative by moving multimodal reasoning from the language space into the visual embedding space \citep{li2025latentvisualreasoning}. 
Instead of generating only textual thoughts, LVR trains the language model to produce latent visual states that reconstruct query-relevant visual tokens and interleaves them with text generation, allowing the model to reason over visual semantics without verbalizing all intermediate evidence. 
While effective for fine-grained visual understanding and perception-intensive question answering, existing latent visual reasoning methods still largely follow a single linear reasoning trajectory \citep{li2025latentvisualreasoning}. 
This formulation can be limiting for questions involving multiple plausible objects, visually similar candidates, or relational evidence across regions, where competing visual hypotheses should be preserved before final prediction \citep{hudson2019gqa,shao2024visualcot}. 
Forcing such alternatives into a single latent path may blur distinct hypotheses and make the learned representation less discriminative.

To address these limitations, we propose \textbf{Latent Visual Contrastive Reasoning} (LVCR), a branch-aware latent reasoning framework for visual question answering.
Instead of following a single latent reasoning trajectory, LVCR represents multiple question-conditioned visual hypotheses in parallel.
Each branch softly attends to image tokens and encodes a candidate evidence state, which is injected into the language model as an additional latent visual reasoning token.
This allows answer generation to condition on multiple evidence configurations within the model's latent space, without external visual tools or repeated image re-encoding.

To make the branch space evidence-aware, we use bounding-box annotations from VISUAL COT as weak supervision \citep{shao2024visualcot}.
We introduce a differentiable bbox-guided contrastive alignment objective \citep{oord2018representation} that pulls the evidence-weighted branch representation toward annotated answer-relevant visual evidence \citep{radford2021learning} while contrasting it against negative visual regions.

Our contributions are summarized as follows:
\begin{itemize}
    \item We develop a latent reasoning framework (LVCR) that represents multiple question-conditioned visual hypotheses instead of a single latent reasoning trajectory.
    
    \item We propose a soft bbox-guided contrastive branch-evidence alignment method that aligns latent branches with answer-relevant visual evidence while contrasting them against negative evidence regions.
    
    \item We introduce a lightweight branch injection and ranking mechanism that feeds multiple latent visual branches into the decoder as reasoning tokens, enabling branch-aware answer generation through question-relevant branch ordering.

    \item We demonstrate the effectiveness of LVCR on perception-intensive visual reasoning benchmarks at both 3B and 7B scales, showing improvements over backbone models and single-path latent reasoning baselines.
\end{itemize}

\section{Related Work}

\subsection{Multimodal LLM for Visual Reasoning}

Multimodal large language models (MLLMs) have become a dominant paradigm for vision-language understanding by connecting visual encoders with large language models. Early representative systems such as Flamingo show that large language models can be adapted to interleaved image-text inputs and perform few-shot vision-language tasks \citep{alayrac2022flamingo}. PaLI further demonstrates the effectiveness of jointly scaling language-image models across multilingual and multitask settings \citep{chen2022pali}. Subsequent work improves multimodal alignment by using frozen visual encoders and lightweight modality bridges \citep{li2023blip2,dai2023instructblip}. Instruction-tuned MLLMs further enhance open-ended visual dialogue and multimodal task following \citep{liu2023llava,liu2024improved}. Recent general-purpose MLLMs further strengthen visual recognition, object localization, document understanding, and multimodal instruction following through improved visual encoders, instruction data, and scaling recipes \citep{bai2025qwen25vl,tong2024cambrian1,deitke2024molmo,chen2024internvl25,zhu2025internvl3}. These advances have enabled strong performance on visual question answering and compositional visual reasoning benchmarks \citep{antol2015vqa,hudson2019gqa}. However, many visual reasoning problems require more than global image understanding: models often need to identify question-relevant evidence, compare visually similar candidates, and reason over fine-grained attributes or spatial relations. Recent studies have also extended multimodal foundation models to specialized scientific and medical reasoning scenarios \cite{zhang2026nanophotonics,tang2026ctdelta}.

\subsection{Chain-of-Thought and Region-Grounded Visual Reasoning}

Chain-of-thought prompting improves complex reasoning in language models by eliciting intermediate textual rationales before producing final answers \citep{wei2022chain}. This idea has also been extended to multimodal settings, where models generate rationales conditioned on both images and text to support answer prediction \citep{zhang2023multimodalcot}. Interactive visual chain-of-thought prompting further decomposes knowledge-based visual reasoning into perception, reasoning, and confirmation stages \citep{chen2024visual}. While such methods improve interpretability and often help reasoning performance, their intermediate reasoning process is primarily represented in the language space.

To better expose visual evidence, another line of work introduces explicit region-grounded, interactive, or tool-augmented reasoning. Visual programming methods compose off-the-shelf vision modules and symbolic operations to solve compositional visual tasks without task-specific training \citep{gupta2022visprog}. Tool-augmented systems use language models to generate programs or invoke external visual experts for multi-step visual reasoning \citep{suris2023vipergpt,lu2023chameleon}. Interactive systems further connect language models with external visual foundation models for perception, editing, and reasoning \citep{wu2023visual,yang2023mm}. In parallel, region-grounded MLLMs support spatial-coordinate inputs and outputs for referential dialogue, localization-aware visual reasoning, and fine-grained grounding \citep{peng2023kosmos,chen2023shikra,you2023ferret,li2024groundinggpt}. Recent spatially grounded VLMs further emphasize region-level and 3D-aware spatial reasoning over objects and relations \citep{chen2024spatialvlm,cheng2024grounded}. Visual CoT provides large-scale multimodal reasoning data with bounding-box evidence and step-by-step rationales, encouraging models to focus on answer-relevant image regions \citep{shao2024visualcot}. These studies demonstrate the importance of grounding reasoning in localized visual evidence, but many of them rely on explicit textual rationales, external visual tools, hand-designed operations, or repeated visual processing.

\subsection{Latent-Space Reasoning and Visual Alignment}

Recent work has begun to investigate whether intermediate reasoning can be represented in continuous latent spaces rather than only as discrete text tokens. In language reasoning, continuous-thought methods reuse hidden states as latent reasoning states instead of verbalizing every intermediate step \citep{hao2024training}. In multimodal reasoning, Latent Visual Reasoning (LVR) proposes to generate latent visual states that reconstruct query-relevant visual tokens, thereby enabling autoregressive reasoning directly in the visual embedding space \citep{li2025latentvisualreasoning,wang2026scdm}. Recent multimodal continuous-thought work further explores latent-space reasoning over visual and textual representations \citep{pham2025multimodal,zheng2025corosam}. This direction is closely related to broader efforts in representation learning and vision-language alignment. Contrastive learning objectives learn representations by pulling positive pairs together and separating them from negative samples \citep{oord2018representation}. Large-scale vision-language pretraining methods further show that contrastive alignment between visual and textual representations can produce transferable multimodal representations \citep{jia2021scaling,radford2021learning,zhai2023sigmoid}. Together, latent-space reasoning and contrastive alignment provide useful foundations for learning visually grounded intermediate representations. However, how to structure and supervise multiple latent visual hypotheses for fine-grained multimodal reasoning remains relatively underexplored.

\begin{figure*}[t]
    \centering
    \includegraphics[width=\textwidth]{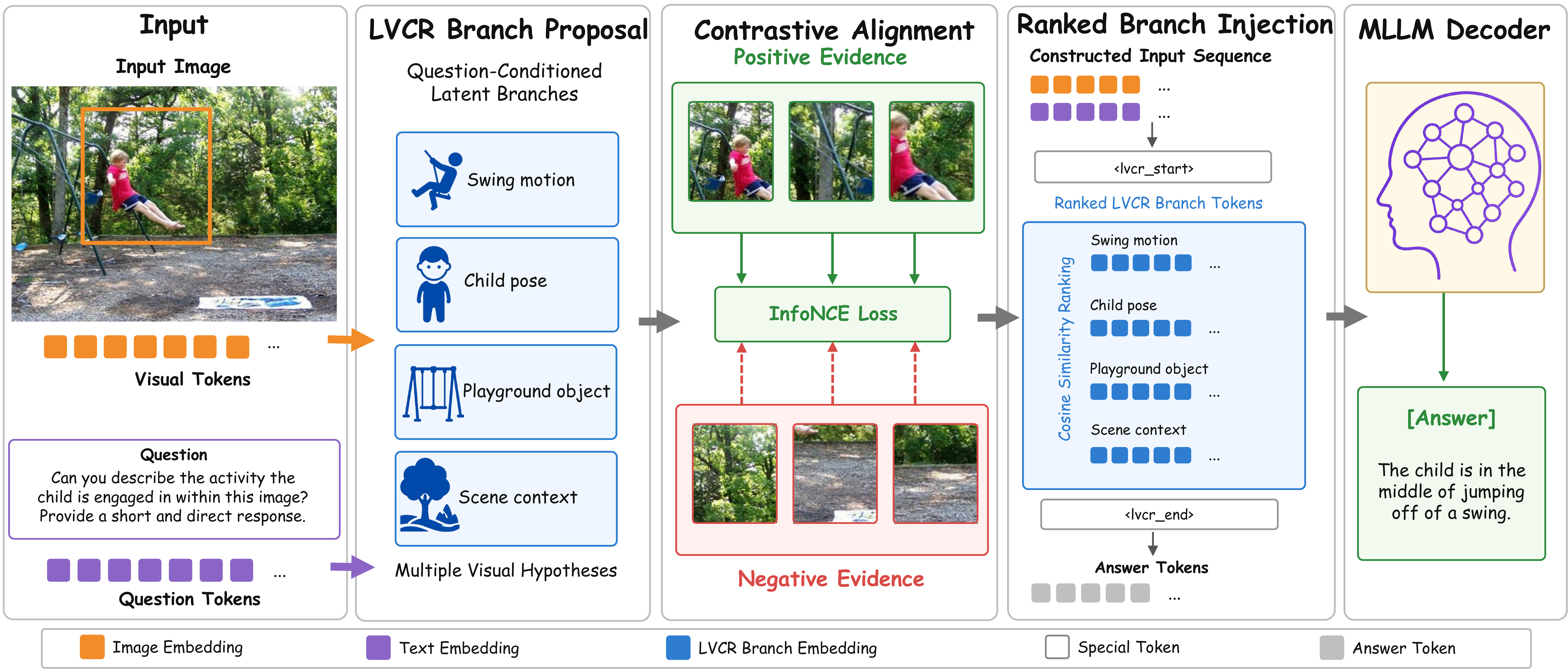}
\caption{
Overall framework of LVCR. Given an image-question pair, LVCR encodes the image and question into visual and textual tokens, and proposes multiple question-conditioned latent branches as candidate visual hypotheses over the image. These branches are contrastively aligned with answer-relevant visual evidence and separated from irrelevant regions. LVCR then ranks the branches using relevance scores computed by cosine similarity with the question representation, and injects the selected branch tokens into the MLLM decoder, which combines them with the original multimodal context to generate the final answer.
}
    \label{fig:framework}
\end{figure*}

\section{Method}

Fig.~\ref{fig:framework} shows the overall pipeline (from left to right) of the proposed LVCR. Given an image--question pair, LVCR first encodes the image into visual tokens and the question into a textual representation using a pretrained vision-language backbone. 
It then constructs $K$ question-conditioned latent branches, where each branch softly attends to the visual tokens and aggregates them into a candidate evidence state, thereby maintaining multiple latent visual hypotheses before answer decoding. 
During training, LVCR uses answer-relevant bounding-box annotations as localized evidence supervision: the annotated region is converted into a positive evidence representation, and a soft contrastive branch-evidence alignment loss pulls evidence-matched branches toward this representation while contrasting them against negative visual regions. 
The ranked branches are finally injected into the language decoder as latent reasoning tokens, and the whole model is optimized with a compact objective combining answer prediction and bbox-guided contrastive evidence alignment. 

\subsection{Question-Conditioned Branch Proposal}

Let \(I\) denote the input image. The vision-language backbone encodes \(I\) into a visual token sequence
\begin{equation}
V=\{v_i\}_{i=1}^{N}, \qquad v_i\in\mathbb{R}^{d},
\end{equation}
and produces a question representation \(q\in\mathbb{R}^{d}\).
LVCR introduces $K$ learnable branch queries $\{e_k\}_{k=1}^{K}$ to parameterize a set of latent visual branches. 
For each branch $k$, we compute a question-conditioned relevance score for each visual token:
\begin{equation}
s_{k,i}=f_{\mathrm{br}}(v_i,q,e_k),
\end{equation}
where $e_k$ is the $k$-th learnable branch query, and $f_{\mathrm{br}}(\cdot)$ is a learnable scoring function used to compute branch-specific attention scores over visual tokens conditioned on the question.
The scores are normalized over the visual tokens to obtain the branch attention weights:
\begin{equation}
a_{k,i}
=
\frac{\exp(s_{k,i})}
{\sum_{j=1}^{N}\exp(s_{k,j})}.
\end{equation}
The latent representation of branch $k$ is then computed as the attention-weighted visual state
\begin{equation}
b_k=\sum_{i=1}^{N}a_{k,i}v_i .
\end{equation}
Collecting all branches gives
\begin{equation}
\mathcal{B}=\{b_k\}_{k=1}^{K}.
\end{equation}
Because the attention weights depend jointly on the visual token $v_i$, the question representation $q$, and the branch query $e_k$, different branches can specialize to different question-conditioned visual evidence patterns while remaining differentiable. 
The next section describes how localized evidence annotations are converted into supervision for these latent branches.

\subsection{BBox-Guided Evidence Representation}
During training, we assume access to answer-relevant bounding-box annotations that indicate the visual evidence required to answer each question. These annotations can come from any dataset or supervision source that provides localized evidence regions, and our formulation is not tied to a specific dataset. We use the annotated regions to construct a target evidence representation for contrastive supervision.

Let $m_i \in \{0,1\}$ indicate whether visual token $v_i$ falls inside the annotated bounding box. 
The target evidence representation is obtained by pooling the visual tokens inside the bounding box:
\begin{equation}
r^+
=
\frac{
\sum_{i=1}^{N} m_i v_i
}{
\sum_{i=1}^{N} m_i
}.
\end{equation}

The representation $r^+$ serves as the positive visual evidence for the current image-question pair. 
Intuitively, a good branch proposal module should produce at least one branch that is strongly aligned with this evidence representation.

\subsection{Contrastive Branch-Evidence Alignment}

The goal of contrastive branch-evidence alignment is to guide latent branches toward answer-relevant visual evidence without forcing the annotated region to be assigned to a single branch.
Since different branches may capture partially relevant evidence, we use a soft evidence-matching mechanism to distribute supervision across all branches.
Given the proposed branches $\mathcal{B}$ and the evidence representation $r^+$, we compute a soft matching weight for each branch:
\begin{equation}
w_k
=
\frac{
\exp(\mathrm{sim}(b_k, r^+) / \tau_s)
}{
\sum_{j=1}^{K}
\exp(\mathrm{sim}(b_j, r^+) / \tau_s)
},
\end{equation}
where $\mathrm{sim}(\cdot,\cdot)$ denotes cosine similarity and $\tau_s$ is a soft assignment temperature.
The weight $w_k$ measures how strongly the $k$-th branch is associated with the annotated evidence, assigning larger weights to evidence-relevant branches while still keeping all branches differentiable.

We then construct a soft positive branch representation by aggregating all candidate branches according to their evidence-matching weights:
\begin{equation}
\bar{b}^{+}
=
\sum_{k=1}^{K}
w_k b_k .
\end{equation}
This soft aggregation avoids a non-differentiable hard branch selection and enables credit assignment across multiple potentially relevant branches.

The contrastive branch-evidence alignment loss is defined as
\begin{equation}
\mathcal{L}_{\mathrm{NCE}}
=
-\log
\frac{
\exp(s^{+}/\tau)
}{
\exp(s^{+}/\tau)
+
\sum_{k=1}^{K}\exp(s_k^{-}/\tau)
},
\end{equation}
where
\begin{equation}
s^{+}=\mathrm{sim}(\bar{b}^{+}, r^+),
\qquad
s_k^{-}=\mathrm{sim}(b_k, r^-).
\end{equation}
Here, $\tau$ is the contrastive temperature and $r^-$ denotes negative evidence representations sampled from low-overlap candidate regions or other images in the mini-batch.

This objective pulls the softly aggregated positive branch representation toward the annotated evidence while contrasting individual branches against negative evidence regions.
As a result, the proposal module learns visually grounded branch representations without requiring an explicit branch selector or a manually designed diversity regularizer.

\subsection{Differentiable Branch Credit Assignment}

The soft aggregation in Eq.~(8) can also be interpreted as a differentiable credit-assignment mechanism over the latent branch set. Let
\begin{equation}
    c_k = \mathrm{sim}(b_k,r^+)
\end{equation}
denote the evidence compatibility of branch $k$. The normalized weight $w_k$ is a softmax function of $c_k$, so every branch participates in the positive representation $\bar{b}^{+}$, but branches that are more compatible with the annotated evidence receive larger coefficients. Consequently, the contrastive objective does not require a discrete branch label or a non-differentiable selection step. Instead, the branch proposal module can gradually redistribute evidence responsibility among the $K$ branches during optimization.

This formulation is useful because the supervision specifies an answer-relevant image region but does not specify how that region should be decomposed across latent hypotheses. A hard assignment would force the training procedure to choose a single branch even when several branches capture complementary parts of the same evidence. In LVCR, the soft assignment preserves this ambiguity: strongly matched branches dominate the evidence-aligned representation, while partially matched branches can still receive learning signals through both their contribution to $\bar{b}^{+}$ and the normalization over branch similarities. At the same time, the negative term in Eq.~(9) discourages branches from indiscriminately matching visually irrelevant regions. The resulting training signal therefore combines positive evidence attraction with branch-level contrast against negative visual content.

\subsection{Answer Prediction}

The proposed branches are injected back into the vision-language model as additional latent visual reasoning tokens.
Concretely, we reserve $K$ special \texttt{<lvcr>} token positions in the input sequence, placed between a \texttt{<lvcr\_start>} marker and the answer.
All $K$ branches are injected and the language model's own multi-head self-attention integrates the branch information without any separate fusion module.
The decoder then predicts the final answer auto-regressively:
\begin{equation}
p(y \mid I, q)
=
\mathrm{Decoder}\!\left(y \;\middle|\; q,\, b_1, \ldots, b_K,\, V\right).
\end{equation}
The answer prediction loss is the standard negative log-likelihood:
\begin{equation}
\mathcal{L}_{\mathrm{ans}}
=
-\sum_{t=1}^{T}
\log p(y_t \mid y_{<t}, I, q).
\end{equation}

At inference time, we rank the $K$ branches by their cosine similarity to the question representation $q$ before injection.
The branches are injected in descending order of similarity, so that the most question-relevant branch occupies the first \texttt{<lvcr>} position.
This ordering prioritizes the strongest branch signal while preserving all branches for complementary visual context.

\subsection{Ranked Latent Sequence Construction}

The branch representations are continuous vectors and therefore do not have an intrinsic left-to-right order. Before they are inserted into the language model at inference time, LVCR converts the unordered branch set into a deterministic latent reasoning sequence. We first compute a question-relevance score for every branch,
\begin{equation}
    \rho_k = \mathrm{sim}(b_k,q),
\end{equation}
and define a permutation $\pi$ such that
\begin{equation}
    \rho_{\pi(1)} \geq \rho_{\pi(2)} \geq \cdots \geq \rho_{\pi(K)}.
\end{equation}
The ordered latent branch sequence is then
\begin{equation}
    B^{\mathrm{rank}} =
    \left[b_{\pi(1)}, b_{\pi(2)}, \ldots, b_{\pi(K)}\right].
\end{equation}
This sequence is placed between the two boundary tokens to form the latent reasoning scaffold
\begin{equation}
\mathcal{S}_{\mathrm{LVCR}}=
\left[
\texttt{<lvcr\_start>},
B^{\mathrm{rank}},
\texttt{<lvcr\_end>}
\right].
\end{equation}
The scaffold is concatenated with the original multimodal context before answer decoding. Importantly, ranking changes only the positions of the branch states; it does not discard any branch. Thus, the first latent position receives the branch most directly aligned with the question, while later positions retain alternative or complementary visual hypotheses. The decoder can attend to all $K$ branch states together with the original image and question tokens, allowing the final prediction to combine the strongest evidence hypothesis with additional contextual cues.

During training, the branch states remain fully differentiable and are optimized jointly through answer prediction and contrastive evidence alignment. During inference, the evidence annotation and the contrastive target are removed, and the question-conditioned ranking provides the only ordering operation. This creates a simple transition from supervised branch organization during training to annotation-free latent reasoning at test time.

\subsection{Training Objective}

We train LVCR to answer the question correctly while guiding its latent branches toward answer-relevant visual evidence.
Accordingly, the final objective combines the standard answer prediction loss with bbox-guided contrastive branch-evidence alignment:
\begin{equation}
\mathcal{L}
=
\mathcal{L}_{\mathrm{ans}}
+
\lambda_{\mathrm{nce}}
\mathcal{L}_{\mathrm{NCE}},
\end{equation}
where $\lambda_{\mathrm{nce}}$ controls the strength of the contrastive alignment objective.

The answer loss remains the primary learning signal, while the NCE loss regularizes the latent branches to capture localized visual evidence related to the answer.
This yields a compact end-to-end training formulation that couples language supervision with visual evidence alignment.

\subsection{Implementation Details}

LVCR is implemented as a lightweight latent-reasoning module on top of the representations already produced by the pretrained multimodal backbone. The visual sequence $V$ and question representation $q$ in Section~III-A share the hidden dimension $d$, and the $K$ branch queries are trainable parameters optimized jointly with the language model and the remaining LVCR parameters. The branch proposal module therefore operates on the encoded visual tokens directly rather than introducing a second visual encoder or an additional image-processing stage.

For localized supervision, each annotated bounding box is mapped to the visual-token grid to construct the binary mask $\{m_i\}_{i=1}^{N}$ used in Eq.~(6). The positive evidence representation is pooled only from the tokens selected by this mask, while negative evidence is obtained from low-overlap candidate regions or from other images in the mini-batch, as described in Section~III-C. The bounding-box annotation is used only to construct the training-time contrastive target; neither bounding boxes nor the contrastive loss are required at inference time.

The latent branches are inserted through the reserved reasoning scaffold $\texttt{<lvcr\_start>}$, $K$ $\texttt{<lvcr>}$ positions, and $\texttt{<lvcr\_end>}$. The branch vectors occupy the $K$ reserved latent positions and are processed together with the original multimodal context by the decoder's existing self-attention layers. Consequently, LVCR does not require an additional branch-fusion network. During training, we optimize the language model and LVCR parameters while keeping the vision encoder and multimodal merger frozen; the numerical optimization settings are given in Section~IV-A.

\begin{algorithm}[t]
\caption{Training and inference procedure of LVCR.}
\label{alg:lvcr}
\small
\begin{algorithmic}[1]
\REQUIRE Image $I$, question $Q$, branch number $K$; training additionally uses bounding box $R$ and answer $y$
\ENSURE Predicted answer $\hat{y}$
\STATE Encode $I$ and $Q$ to obtain visual tokens $V=\{v_i\}_{i=1}^{N}$ and question representation $q$
\FOR{$k=1,\ldots,K$}
    \STATE Compute $s_{k,i}=f_{\mathrm{br}}(v_i,q,e_k)$ over all visual tokens
    \STATE Normalize $\{s_{k,i}\}_{i=1}^{N}$ and aggregate $b_k=\sum_i a_{k,i}v_i$
\ENDFOR
\IF{training}
    \STATE Map $R$ to the visual-token mask $\{m_i\}$ and construct positive evidence $r^+$
    \STATE Sample negative evidence $r^-$ and compute soft branch weights $\{w_k\}$
    \STATE Form $\bar{b}^{+}=\sum_k w_k b_k$ and compute $\mathcal{L}_{\mathrm{NCE}}$
    \STATE Insert $\{b_k\}$ into the reserved $\texttt{<lvcr>}$ positions and compute $\mathcal{L}_{\mathrm{ans}}$
    \STATE Update trainable parameters using $\mathcal{L}_{\mathrm{ans}}+\lambda_{\mathrm{nce}}\mathcal{L}_{\mathrm{NCE}}$
\ELSE
    \STATE Rank branches by $\mathrm{sim}(b_k,q)$ in descending order
    \STATE Insert all ranked branches into the reserved $\texttt{<lvcr>}$ positions
    \STATE Autoregressively decode the answer $\hat{y}$
\ENDIF
\end{algorithmic}
\end{algorithm}

Algorithm~\ref{alg:lvcr} summarizes the complete training and inference paths. Training uses localized evidence only to shape the latent branch space, whereas inference constructs, ranks, and injects the branches from the image--question pair alone. This separation is important because evidence supervision improves the internal reasoning representation without introducing an annotation requirement at deployment time.

\subsection{Computational Complexity and Efficiency}

LVCR adds computation primarily through branch proposal and the insertion of a small number of latent tokens. Let $N$ denote the number of visual tokens, $d$ the hidden dimension, $K$ the number of branches, and $L$ the sequence length seen by the decoder before LVCR tokens are inserted. Evaluating the branch scoring function over all branch--token pairs and aggregating the branch representations requires $O(KNd)$ computation when the scoring operation is linear in $d$, together with $O(KN)$ attention weights. The soft branch-evidence assignment and contrastive similarities add only $O(Kd)$ computation once the positive and negative evidence representations have been formed. Ranking the branches at inference time costs $O(K\log K)$.

For answer generation, LVCR increases the decoder sequence length from $L$ to $L+K$. Under standard self-attention, the corresponding per-layer attention term changes from $O(L^2d)$ to $O((L+K)^2d)$, giving an additional $O((2LK+K^2)d)$ attention term. In our configurations, $K$ is small ($K=4$ for 3B and $K=8$ for 7B), so the number of added latent positions is fixed and much smaller than the full multimodal context. More importantly, all branches are constructed from the visual tokens produced by a single image encoding. LVCR therefore avoids iterative cropping, external visual-tool calls, and repeated vision-encoder passes. The bounding-box pooling and NCE computation are training-only operations and introduce no inference-time annotation or evidence-processing cost. This analysis characterizes the structural overhead of LVCR rather than claiming a particular wall-clock speedup.

\subsection{Design Discussion}

\paragraph{Why multiple latent branches?}
A single latent state must compress all question-relevant visual information into one representation. This is potentially restrictive when the answer depends on several plausible objects, multiple spatial regions, or competing fine-grained cues. LVCR instead maintains a small set of branch states before decoding. Because each branch has its own learnable query and question-conditioned attention distribution, the model can represent different candidate evidence configurations in parallel rather than averaging them immediately into a single trajectory. The decoder can then integrate complementary branch information jointly with the original visual and textual context.

\paragraph{Why soft evidence assignment?}
The bounding box specifies which visual region is relevant to the answer, but it does not provide a semantic label indicating which latent branch should explain that region. A hard one-to-one assignment would therefore introduce an arbitrary branch identity and would discard useful gradients for other partially aligned branches. The soft weights in Eq.~(7) treat branch identity as latent: strongly aligned branches receive larger supervision while all branches remain differentiable. This lets localized evidence organize the branch space without requiring a separate branch selector or a manually specified matching rule.

\paragraph{Why rank before injection?}
The set of candidate branches has no inherent semantic order, whereas the injected $\texttt{<lvcr>}$ positions form an ordered sequence for the language model. Ranking by cosine similarity with the question representation provides a deterministic, question-dependent ordering that places the most directly relevant branch first while retaining the remaining branches as complementary context. This design avoids the information bottleneck of selecting only one hypothesis and removes the permutation ambiguity of injecting an unordered branch set. Section~IV-C evaluates these alternatives and shows that ranked multi-branch injection is more effective than either top-1 or unranked injection.

\section{Experiments}

\subsection{Experimental Setup}

\paragraph{Training data.}
We use VISUAL COT as the primary training data \citep{shao2024visualcot}. 
This large-scale VQA dataset contains question-answer examples with bounding-box evidence annotations. 
After formatting and filtering, our training set contains 417,453 image-question-answer examples. 
Each example contains one annotated bounding box, which we use to construct the positive visual evidence representation for the contrastive branch-evidence alignment loss.

\paragraph{Model and training.}
We build LVCR on top of Qwen2.5-VL-Instruct backbones and report results for both the 3B and 7B variants \citep{bai2025qwen25vl}. 
For 3B models, LVCR uses $K=4$ latent branches, while for 7B models it uses $K=8$; in both cases, the number of inserted \texttt{<lvcr>} tokens is set equal to $K$.
The answer prediction loss is optimized jointly with the contrastive branch-evidence alignment loss using $\lambda_{\mathrm{nce}}=0.1$, contrastive temperature $\tau=0.07$, and soft assignment temperature $\tau_s=0.2$. 
We fine-tune the language model and LVCR parameters while keeping the vision encoder and multimodal merger frozen. 
Training uses bfloat16 precision, gradient checkpointing, DeepSpeed ZeRO-2, AdamW with learning rate $1\times 10^{-5}$, weight decay 0.1, warmup ratio 0.03, and a cosine learning-rate schedule. 
We train the 3B models on two NVIDIA A100 80GB GPUs and the 7B models on four NVIDIA A100 80GB GPUs.

\paragraph{Inference.}
At inference time, we insert the LVCR latent reasoning scaffold before answer generation using a \texttt{<lvcr\_start>} token, $K$ \texttt{<lvcr>} tokens, and a \texttt{<lvcr\_end>} token. 
Before injection, we rank all $K$ branches by their cosine similarity to the question representation and place the most question-relevant branch in the first \texttt{<lvcr>} position. 
All ranked branches are therefore available to the decoder before it emits the final textual answer. 
We also evaluate ablations without branch ranking and with top-1 injection in Section~\ref{para:inference_strategy}, where top-1 denotes using only the highest-ranked branch under the same cosine-similarity criterion. 
All benchmark evaluations use deterministic decoding.

\paragraph{Evaluation benchmarks.}
We evaluate LVCR on perception-intensive multiple-choice visual question answering benchmarks, including $V^*$ Bench, MMVP, and BLINK. 
$V^*$ Bench evaluates fine-grained visual detail search and relative spatial reasoning, and we report the overall score ($V^*$) as well as its Direct Attributes ($V^*_{\mathrm{D.A.}}$) and Relative Position ($V^*_{\mathrm{R.P.}}$) subsets \citep{wu2024vstar}. 
MMVP evaluates whether MLLMs can distinguish fine-grained visual differences under subtle image perturbations \citep{tong2024eyes}. 
For BLINK, we report results on Counting, IQ Test, Jigsaw, Relative Reflectance, and Spatial Relation, which cover perception-intensive skills such as counting, abstract visual reasoning, reflectance comparison, and spatial understanding \citep{fu2024blink}. 
For all benchmarks, the model is prompted to answer directly with the option letter, and accuracy is computed by exact matching against the ground-truth option after standard answer extraction. 
We compare LVCR with the corresponding Qwen2.5-VL-Instruct backbones and additional multimodal reasoning baselines under the same evaluation protocol.

\paragraph{Baselines.}
We compare LVCR with baselines at both 3B and 7B scales. 
At the 3B scale, we include Qwen2.5-VL-3B as the backbone baseline \citep{bai2025qwen25vl}, Vision-R1 as a recent visual reasoning baseline \citep{huang2025visionr1}, a baseline trained with standard supervised fine-tuning (SFT) on the same instruction-format data, denoted as Vanilla SFT, and LVR as the single-path latent visual reasoning baseline \citep{li2025latentvisualreasoning}. 
At the 7B scale, we compare against Qwen2.5-VL-7B, PAPO \citep{wang2025papo}, Vision-R1, PixelReasoner \citep{su2025pixelreasoner}, Vanilla SFT, and LVR \citep{li2025latentvisualreasoning}.  
PAPO, Vision-R1, and PixelReasoner provide comparisons with recent multimodal reasoning methods. 
All methods are evaluated using the same prompting, deterministic decoding, and answer-extraction protocol.

\begin{table*}[t]
\centering
\small
\begin{tabular}{lccccccccc}
\hline
Model & $V^*$ & $V^*_{\mathrm{D.A.}}$ & $V^*_{\mathrm{R.P.}}$ & MMVP & Counting & IQ-Test & Jigsaw & Relative Reflect & Spatial Relation \\
\hline
Qwen2.5-VL-3B & \underline{73.8} & \underline{80.9} & 63.2 & 60.0 & \underline{65.8} & 24.7 & \underline{53.3} & \textbf{42.5} & \underline{81.1} \\
Vision-R1 & 71.7 & 72.2 & \textbf{71.1} & \textbf{66.0} & 60.0 & 20.0 & 44.7 & 32.8 & 76.2 \\
Vanilla SFT & 70.7 & 80.0 & 56.6 & 59.0 & \textbf{67.5} & 22.0 & 52.0 & \textbf{42.5} & \textbf{84.6} \\
LVR & 70.7 & 79.1 & 57.9 & 60.7 & 64.2 & \textbf{28.7} & 52.7 & \underline{40.3} & \underline{81.1} \\
LVCR (Ours) & \textbf{75.4} & \textbf{82.6} & \underline{64.5} & \underline{61.0} & 62.5 & \underline{28.0} & \textbf{54.0} & \textbf{42.5} & \textbf{84.6} \\
\hline
\end{tabular}
\caption{
Main results for 3B-scale models. All numbers are accuracy (\%). 
Vision-R1 and LVR are reproduced on the 3B backbone using their official implementations. 
The best result in each column is shown in bold, and the second-best result is underlined.
}
\label{tab:main_results_3b}
\end{table*}

\begin{table*}[t]
\centering
\small
\begin{tabular}{lccccccccc}
\hline
Model & $V^*$ & $V^*_{\mathrm{D.A.}}$ & $V^*_{\mathrm{R.P.}}$ & MMVP & Counting & IQ-Test & Jigsaw & Relative Reflect & Spatial Relation \\
\hline
Qwen2.5-VL-7B & 78.5 & 81.7 & 73.7 & 66.7 & 66.7 & 26.0 & 52.0 & 38.8 & 87.4 \\
PAPO & 36.1 & 25.2 & 52.6 & 54.3 & 66.7 & \textbf{29.3} & 52.0 & 39.6 & \textbf{88.8} \\
Vision-R1 & 70.2 & 70.4 & 69.7 & 46.7 & 51.7 & 26.7 & 27.3 & \textbf{44.8} & 66.4 \\
PixelReasoner & \textbf{80.1} & 81.7 & \textbf{77.6} & 67.0 & 65.0 & 23.3 & \underline{52.7} & 40.3 & \underline{88.1} \\
Vanilla SFT & 79.1 & \textbf{82.6} & 73.7 & 65.7 & \underline{67.5} & 26.7 & 45.3 & 33.6 & \textbf{88.8} \\
LVR & \underline{79.8} & \underline{82.3} & 76.1 & \underline{70.5} & 67.3 & \underline{27.7} & \underline{52.7} & \underline{42.5} & 86.2 \\
LVCR (Ours) & 79.6 & \textbf{82.6} & \underline{76.3} & \textbf{72.3} & \textbf{70.0} & \textbf{29.3} & \textbf{54.7} & \underline{42.5} & \underline{88.1} \\
\hline
\end{tabular}
\caption{
Main results for 7B-scale models. All numbers are accuracy (\%). 
For PAPO, Vision-R1, and PixelReasoner, we evaluate their publicly released 7B checkpoints. 
For LVR, we retrain the model following its official training recipe. 
The best result in each column is shown in bold, and the second-best result is underlined.
}
\label{tab:main_results_7b}
\end{table*}

\subsection{Main Results and Analysis}

Tables~\ref{tab:main_results_3b} and~\ref{tab:main_results_7b} report the main results at the 3B and 7B scales. 
Overall, LVCR achieves competitive performance across the evaluated perception-intensive visual reasoning benchmarks. 
At the 3B scale, LVCR obtains the best or tied-best results on five out of nine metrics, including the overall $V^*$ score, $V^*_{\mathrm{D.A.}}$, Jigsaw, Relative Reflectance, and Spatial Relation. 
Compared with the Qwen2.5-VL-3B backbone, LVCR improves the overall $V^*$ score and both $V^*$ subsets, suggesting that branch-aware latent reasoning is beneficial for fine-grained visual search and attribute-level perception. 
Compared with the single-path LVR baseline, LVCR improves most metrics, especially on $V^*$, $V^*_{\mathrm{R.P.}}$, Jigsaw, and Spatial Relation. 
This indicates that representing multiple question-conditioned visual hypotheses can provide more useful region-level evidence than relying on a single latent reasoning trajectory.

At the 7B scale, LVCR remains strong against a broader set of baselines, including PAPO, Vision-R1, PixelReasoner, Vanilla SFT, and LVR. 
LVCR achieves the best or tied-best results on $V^*_{\mathrm{D.A.}}$, MMVP, Counting, IQ-Test, and Jigsaw, and obtains second-best performance on several other subsets. 
The gains over the Qwen2.5-VL-7B backbone are particularly clear on MMVP and BLINK subsets, where the model needs to distinguish subtle visual differences, count multiple objects, or reason over spatial and perceptual relations. 
Compared with LVR, LVCR further improves performance on MMVP, Counting, IQ-Test, Jigsaw, and Spatial Relation, showing that the proposed multi-branch formulation brings consistent benefits beyond single-path latent visual reasoning. 
These results support the motivation that complex visual reasoning often involves multiple plausible evidence regions, and that preserving several latent visual hypotheses before answer generation can improve perception-intensive reasoning.

The comparison with Vanilla SFT also suggests that the improvements do not simply come from additional supervised fine-tuning. 
Although Vanilla SFT improves some individual metrics, its performance is less stable across benchmarks and model scales. 
For example, at the 3B scale, Vanilla SFT improves Spatial Relation but underperforms the backbone on the overall $V^*$ score and relative-position reasoning. 
This suggests that standard supervised fine-tuning alone may not reliably improve fine-grained visual reasoning. 
In contrast, LVCR combines answer supervision with bbox-guided contrastive branch-evidence alignment, which encourages the latent branches to capture answer-relevant visual regions while still supporting final answer prediction.

The results also reveal that LVCR is not uniformly best on every benchmark. 
Vision-R1 performs strongly on MMVP and $V^*_{\mathrm{R.P.}}$ at the 3B scale, while PixelReasoner achieves the best overall $V^*$ and $V^*_{\mathrm{R.P.}}$ results at the 7B scale. 
These cases suggest that different reasoning methods may emphasize different visual capabilities, such as pixel-level reasoning, visual robustness, or spatial localization. 
Nevertheless, LVCR provides broad improvements over the backbone and single-path latent reasoning baselines across many perception-intensive subsets. 
This demonstrates that branch-aware latent visual reasoning is a useful and complementary mechanism for improving multimodal visual reasoning.

\begin{figure}[t]
    \centering
    \includegraphics[width=0.9\linewidth]{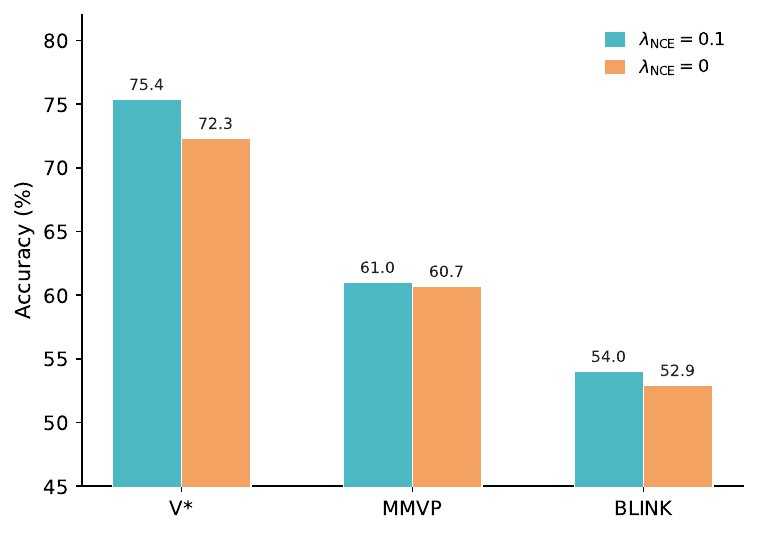}
    \caption{
    Ablation study on the NCE loss coefficient using LVCR-3B with $K{=}4$.
    We compare the default setting with $\lambda_{\mathrm{NCE}}=0.1$ against a variant trained without the NCE loss, i.e., $\lambda_{\mathrm{NCE}}=0$.
    Removing the NCE loss leads to consistent performance drops on $V^*$, MMVP, and BLINK, indicating that contrastive branch-evidence alignment helps latent branches capture question-relevant visual cues.
    }
    \label{fig:nce_ablation}
\end{figure}

\begin{figure}[t]
    \centering
    \includegraphics[width=0.95\linewidth]{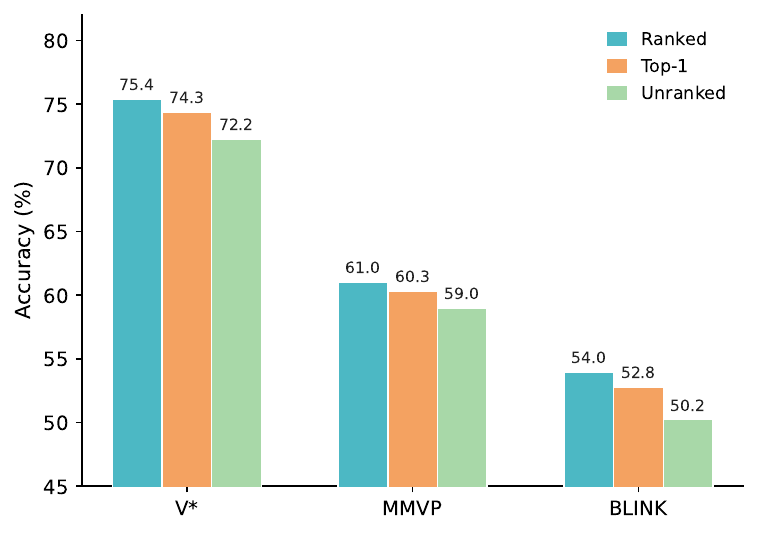}
    \caption{Accuracy (\%) of three branch injection strategies on $V^*$, MMVP, and BLINK (aggregated). \textit{Ranked} denotes our default LVCR strategy; \textit{Top-1} injects only the highest-ranked branch; \textit{Unranked} injects all branches without ordering. Results are reported using LVCR-3B with $K{=}4$.}
    \label{fig:inference_ablation}
\end{figure}

\subsection{Ablation Studies}

\paragraph{Ablation on contrastive alignment.}
We first study the effect of the contrastive branch-evidence alignment objective.
To isolate its contribution, we retrain LVCR-3B by setting the coefficient of the NCE loss to zero, while keeping the remaining training configuration unchanged.
As shown in Figure~\ref{fig:nce_ablation}, removing the NCE loss consistently degrades performance across the evaluated benchmarks.
In particular, the model without NCE loss drops from 75.4 to 72.3 on $V^*$ and from 54.0 to 52.9 on BLINK, while showing a smaller decrease on MMVP.

These results suggest that the NCE loss provides useful supervision for organizing latent branches around question-relevant visual evidence.
Without contrastive branch-evidence alignment, different latent branches are still trained through the answer prediction objective, but they receive weaker guidance on which visual regions or hypotheses they should specialize in.
As a result, the learned branches may become less discriminative and less consistently aligned with the visual cues required by the question.
The improvement brought by the NCE loss is especially visible on $V^*$ and BLINK, which require fine-grained visual grounding and compositional perception.
This supports our design choice of using contrastive supervision to encourage more informative and evidence-aware latent visual branches.

\paragraph{Inference strategy.}
\label{para:inference_strategy}
We ablate how latent branches are organized during inference.
Our default LVCR strategy ranks all $K$ branches by their cosine similarity to the question representation and injects them in descending order of relevance.
We compare it with two alternatives: injecting only the top-1 ranked branch (\textit{Top-1}) and injecting all branches without ranking (\textit{Unranked}).

As shown in Figure~\ref{fig:inference_ablation}, \textit{Ranked} consistently achieves the best performance across $V^*$, MMVP, and BLINK.
On $V^*$, \textit{Ranked} obtains 75.4\% accuracy, outperforming \textit{Top-1} by 1.1 points and \textit{Unranked} by 3.2 points.
On BLINK, the improvement is more pronounced, with \textit{Ranked} exceeding \textit{Top-1} and \textit{Unranked} by 1.2 and 3.8 points, respectively.
The gap between \textit{Ranked} and \textit{Top-1} suggests that relying on a single branch creates an information bottleneck, as complementary visual cues from other branches are discarded.
Meanwhile, the weaker performance of \textit{Unranked} despite using all branches indicates that relevance-based ordering provides useful structural guidance for latent visual reasoning.
These results motivate ranked multi-branch injection as the default inference strategy.

\section{Conclusion}

In this paper, we proposed LVCR, a branch-aware latent reasoning framework for perception-intensive multimodal reasoning. LVCR generates multiple latent visual branches and uses bounding-box supervision with a soft contrastive objective to align each branch with answer-relevant visual evidence. At inference time, LVCR ranks these branches by question relevance and injects the selected latent tokens into the decoder for answer prediction. Experiments at both 3B and 7B scales show that LVCR achieves consistent improvements over competitive baselines. These results suggest that structured multi-branch latent reasoning is a promising direction for visually grounded latent-space reasoning.

\bibliographystyle{IEEEtran}
\bibliography{custom}

% \appendix

% \section{Example Appendix}
% \label{sec:appendix}

% This is an appendix.

\end{document}